\documentclass[hidelinks]{article}
\usepackage{ismir,amsmath,cite,url}
\usepackage{graphicx}
\usepackage{arydshln}
\usepackage{booktabs}
\usepackage{multirow}
\usepackage{amsmath}
\usepackage{appendix}
\graphicspath{{figs/}}
\usepackage{enumitem}
\setitemize{noitemsep,topsep=0pt,parsep=0pt,partopsep=0pt}

\newcommand{\appref}[1]{Appendix~\ref{#1}}

\title{Convolutional Generative Adversarial Networks with Binary Neurons for Polyphonic Music Generation}

\oneauthor
 {Hao-Wen Dong and Yi-Hsuan Yang}
 {Research Center for IT innovation, Academia Sinica, Taipei, Taiwan\\ {\normalsize \tt \{salu133445,yang\}@citi.sinica.edu.tw}}

\def\z{\mathbf{z}}
\def\x{\mathbf{x}}
\def\xh{\mathbf{\hat{x}}}
\def\E{\mathbf{E}}

\begin{document}

\maketitle

\begin{abstract}
It has been shown recently that deep convolutional generative adversarial networks (GANs) can learn to generate music in the form of piano-rolls, which represent music by binary-valued time-pitch matrices. However, existing models can only generate real-valued piano-rolls and require further post-processing, such as hard thresholding (HT) or Bernoulli sampling (BS), to obtain the final binary-valued results. In this paper, we study whether we can have a convolutional GAN model that directly creates binary-valued piano-rolls by using binary neurons. Specifically, we propose to append to the generator an additional refiner network, which uses binary neurons at the output layer. The whole network is trained in two stages. Firstly, the generator and the discriminator are pretrained. Then, the refiner network is trained along with the discriminator to learn to binarize the real-valued piano-rolls the pretrained generator creates. Experimental results show that using binary neurons instead of HT or BS indeed leads to better results in a number of objective measures. Moreover, deterministic binary neurons perform better than stochastic ones in both objective measures and a subjective test. The source code, training data and audio examples of the generated results can be found at \url{https://salu133445.github.io/bmusegan/}.
\end{abstract}

\section{Introduction}
\label{sec:introduction}

\begin{figure}[t]
\centering
\begin{minipage}{0.05\linewidth}
  \raggedleft
  Dr. \\[3.2pt]
  Pi. \\[3.2pt]
  Gu. \\[3.2pt]
  Ba. \\[3.2pt]
  En. \\[3.2pt]
  Re. \\[3.2pt]
  S.L.\\[3.2pt]
  S.P.\\[4.5pt]
  Dr. \\[3.2pt]
  Pi. \\[3.2pt]
  Gu. \\[3.2pt]
  Ba. \\[3.2pt]
  En. \\[3.2pt]
  Re. \\[3.2pt]
  S.L.\\[3.2pt]
  S.P.
\end{minipage}
\hspace*{\fill}
\begin{minipage}{0.91\linewidth}
  \includegraphics[width=\linewidth]{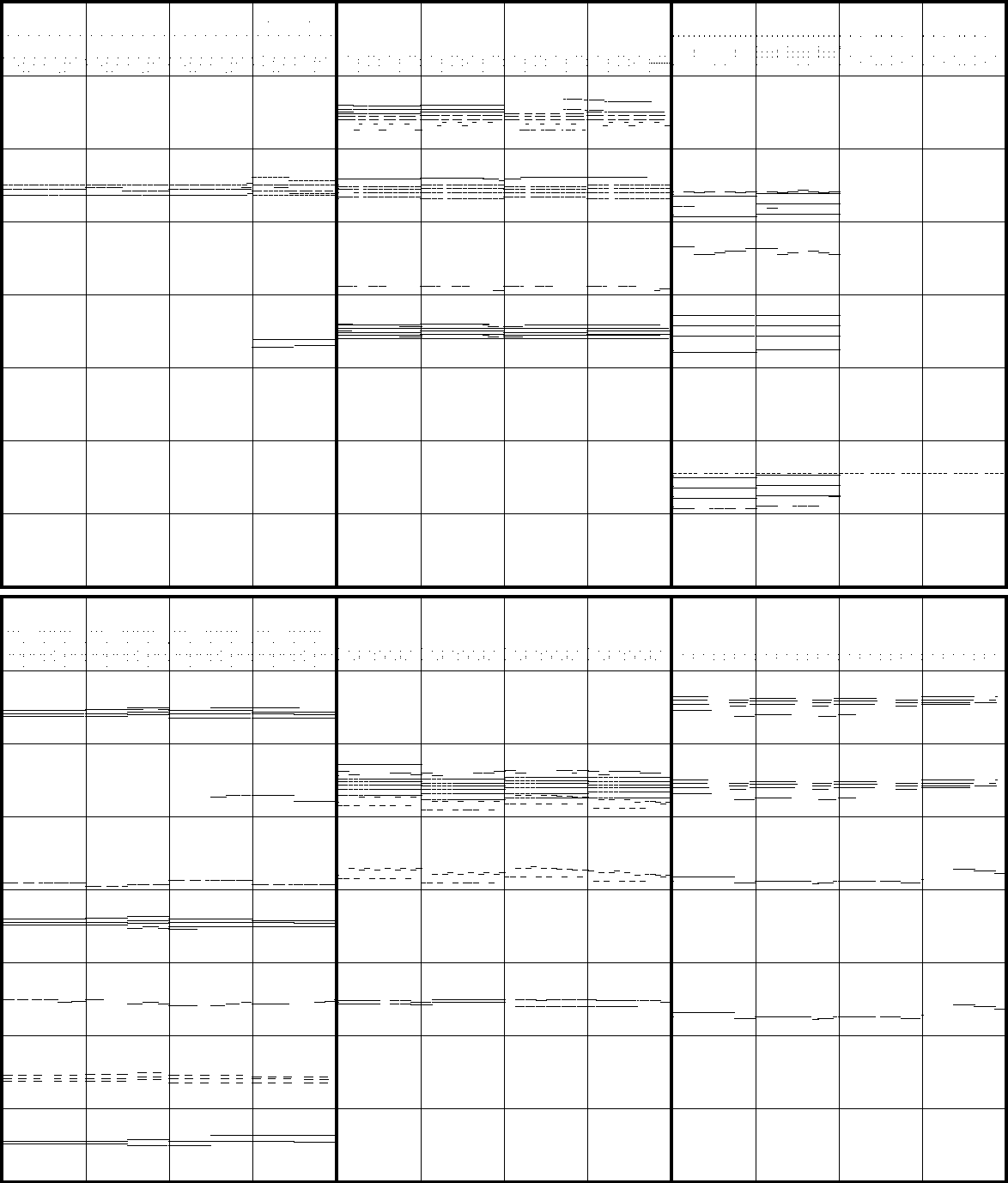}
\end{minipage}
\caption{Six examples of eight-track piano-roll of four-bar long (each block represents a bar) seen in our training data. The vertical and horizontal axes represent note pitch and time, respectively. The eight tracks are \textit{Drums, Piano, Guitar, Bass, Ensemble, Reed, Synth Lead} and \textit{Synth Pad}.} 
\label{fig:training_data}
\end{figure}

Recent years have seen increasing research on symbolic-domain music generation and composition using deep neural networks \cite{briot17survey}. Notable progress has been made to generate monophonic melodies \cite{folkrnn,seqgan}, lead sheets (i.e., melody and chords) \cite{lstm_improvisation,song_from_pi,midinet}, or four-part chorales \cite{deepbach}. To add something new to the table and to increase the polyphony and the number of instruments of the generated music, we attempt to generate piano-rolls in this paper, a music representation that is more general (e.g., comparing to leadsheets) yet less studied in recent work on music generation.  As \figref{fig:training_data} shows, we can consider an $M$-track piano-roll as a collection of $M$ binary time-pitch matrices indicating the presence of pitches per time step for each track.

Generating piano-rolls is challenging because of the large number of possible active notes per time step and the involvement of multiple instruments. Unlike a melody or a chord progression, which can be viewed as a sequence of note/chord events and be modeled by a recurrent neural network (RNN) \cite{lim17ismir,musicVAE}, the musical texture in a piano-roll is much more complicated (see \figref{fig:training_data}). While RNNs are good at learning the temporal dependency of music, convolutional neural networks (CNNs) are usually considered better at learning local patterns \cite{huang17ismir}.

For this reason, in our previous work~\cite{musegan}, we used a convolutional generative adversarial network (GAN)~\cite{gan} to learn to generate piano-rolls of five tracks. We showed that the model generates music that exhibit drum patterns and plausible note events. However, musically the generated result is still far from satisfying to human ears, scoring around $3$ on average on a five-level Likert scale in overall quality in a user study~\cite{musegan}.\footnote{Another related work on generating piano-rolls, as presented by Boulanger-Lewandowski \emph{et al.}~\cite{rbm}, replaced the output layer of an RNN with conditional restricted Boltzmann machines (RBMs) to model high-dimensional sequences and applied the model to generate piano-rolls sequentially (i.e. one time step after another).}

There are several ways to improve upon this prior work. The major topic we are interested in is the introduction of the \emph{binary neurons} (BNs)~\cite{r2rt,bengio} to the model. We note that conventional CNN designs, also the one adopted in our previous work~\cite{musegan}, can only generate real-valued predictions and require further postprocessing \emph{at test time} to obtain the final binary-valued piano-rolls.\footnote{Such binarization is typically not needed for an RNN or an RBM in polyphonic music generation, since an RNN is usually used to predict pre-defined note events~\cite{crnn} and an RBM is often used with binary visible and hidden units and sampled by Gibbs sampling~\cite{rbm,lattner16crbm}.} This can be done by either applying a \emph{hard threshold} (HT) on the real-valued predictions to binarize them (which was done in \cite{musegan}), or by treating the real-valued predictions as probabilities and performing \emph{Bernoulli sampling} (BS).

However, we note that such na\"ive methods for binarizing a piano-roll can easily lead to \textit{overly-fragmented notes}. For HT, this happens when the original real-valued piano-roll has many entries with values close to the threshold. For BS, even an entry with low probability can take the value 1, due to the stochastic nature of probabilistic sampling.

\begin{figure}[t]
\centering
\begin{minipage}{0.45\linewidth}
\centering
\includegraphics[width=\linewidth]{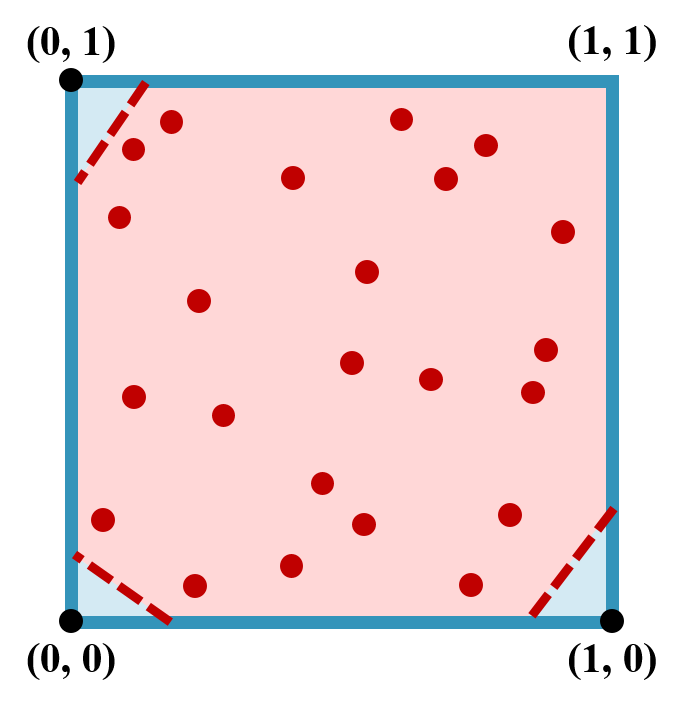}
\end{minipage}
\hspace*{\fill}
\begin{minipage}{0.45\linewidth}
\centering
\includegraphics[width=\linewidth]{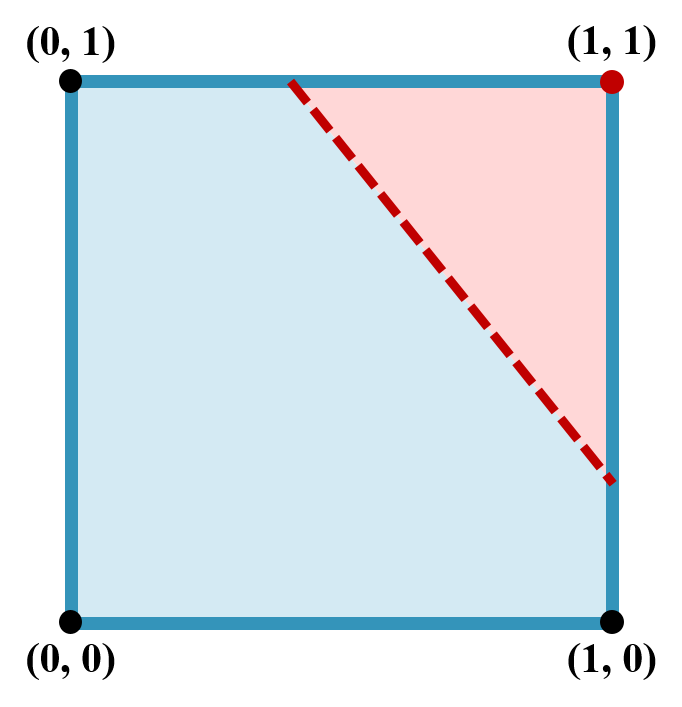}
\end{minipage}
\caption{An illustration of the decision boundaries (red dashed lines) that the discriminator $D$ has to learn when the generator $G$ outputs (left) real values and (right) binary values. The decision boundaries divide the space into the \textit{real} class (in blue) and the \textit{fake} class (in red). The black and red dots represent the real data and the fake ones generated by the generator, respectively. We can see that the decision boundaries are easier to learn when the generator outputs binary values rather than real values.}
\label{fig:theory}
\end{figure}

The use of BNs can mitigate the aforementioned issue, since the binarization is part of the training process. Moreover, it has two potential benefits:
\begin{itemize}
  \item In~\cite{musegan}, binarization of the output of the generator $G$ in GAN is done only at test time not at training time (see \secref{sec:gan} for a brief introduction of GAN). This makes it easy for the discriminator $D$ in GAN to distinguish between the generated piano-rolls (which are real-valued in this case) and the real piano-rolls (which are binary). With BNs, the binarization is done at training time as well, so $D$ can focus on extracting musically relevant features.
  \item Due to BNs, the input to the discriminator $D$ in GAN at training time is binary instead of real-valued. This effectively reduces the model space from $\Re^N$ to $2^N$, where $N$ is the product of the number of time steps and the number of possible pitches. Training $D$ may be easier as the model space is substantially smaller, as \figref{fig:theory} illustrates.
\end{itemize}

Specifically, we propose to append to the end of $G$ a \emph{refiner network} $R$ that uses either deterministic BNs (DBNs) or stocahstic BNs (SBNs) at the output layer. 
In this way, $G$ makes real-valued predictions and $R$ binarizes them. We train the whole network in two stages: in the first stage we pretrain $G$ and $D$ and then fix $G$; in the second stage, we train $R$ and fine-tune $D$. We use residual blocks~\cite{resunit} in $R$ to make this two-stage training feasible (see Section \ref{subsec:refiner}).

As minor contributions, we use a new shared/private design of $G$ and $D$ that cannot be found in~\cite{musegan}. Moreover, we add to $D$ two streams of layers that provide onset/offset and chroma information (see Sections \ref{subsec:generator} and \ref{subsec:discriminator}).

The proposed model is able to directly generate binary-valued piano-rolls at test time. Our analysis shows that the generated results of our model with DBNs features fewer overly-fragmented notes as compared with the result of using HT or BS. Experimental results also show the effectiveness of the proposed two-stage training strategy compared to either a joint or an end-to-end training strategy.

\section{Background}
\label{sec:background}

\subsection{Generative Adversarial Networks}
\label{sec:gan}

A generative adversarial network (GAN)~\cite{gan} has two core components: a \emph{generator} $G$ and a \emph{discriminator} $D$. The former takes as input a random vector $\z$ sampled from a prior distribution $p_\z$ and generates a fake sample $G(\z)$. $D$ takes as input either real data $\x$ or fake data generated by $G$. During training time, $D$ learns to distinguish the fake samples from the real ones, whereas $G$ learns to fool $D$.

An alternative form called WGAN was later proposed with the intuition to estimate the Wasserstein distance between the real and the model distributions by a deep neural network and use it as a critic for the generator~\cite{wgan}. The objective function for WGAN can be formulated as:
\begin{equation} 
\label{eq:wgan}
\min_{G} \max_{D}  \E_{\mathbf{x}\sim p_d}[D(\mathbf{x})] - \E_{\z\sim p_\z}[D(G(\z))]\,,
\end{equation} 
where $p_d$ denotes the real data distribution. In order to enforce Lipschitz constraints on the discriminator, which is required in the training of WGAN, Gulrajani \emph{et al.}~\cite{wgan-gp} proposed to add to the objective function of $D$ a \textit{gradient penalty} (GP) term: $\E_{\hat{\x}\sim p_{\hat{\x}}}[(\nabla_{\hat{\x}}\|\hat{\x}\|-1)^2]$, where $p_{\hat{\x}}$ is defined as sampling uniformly along straight lines between pairs of points sampled from $p_d$ and the model distribution $p_g$. Empirically they found it stabilizes the training and alleviates the mode collapse issue, compared to the weight clipping strategy used in the original WGAN. Hence, we employ WGAN-GP~\cite{wgan-gp} as our generative framework.

\subsection{Stochastic and Deterministic Binary Neurons}

Binary neurons (BNs) are neurons that output binary-valued predictions. In this work, we consider two types of BNs: deterministic binary neurons (DBNs) and stochastic binary neurons (SBNs). DBNs act like neurons with \emph{hard thresholding} functions as their activation functions. We define the output of a DBN for a real-valued input $x$ as:
\begin{equation}
DBN(x) = u(\sigma(x)-0.5)\,,
\label{eq:dbn}
\end{equation}
where $u(\cdot)$ denotes the unit step function and $\sigma(\cdot)$ is the logistic sigmoid function. SBNs, in contrast, binarize an input $x$ according to a probability, defined as:
\begin{equation}
SBN(x) = u(\sigma(x) - v),\; v\sim U[0, 1]\,,
\label{eq:sbn}
\end{equation}
where $U[0, 1]$ denotes a uniform distribution.

\begin{figure}[t]
\centering
\includegraphics[width=\linewidth]{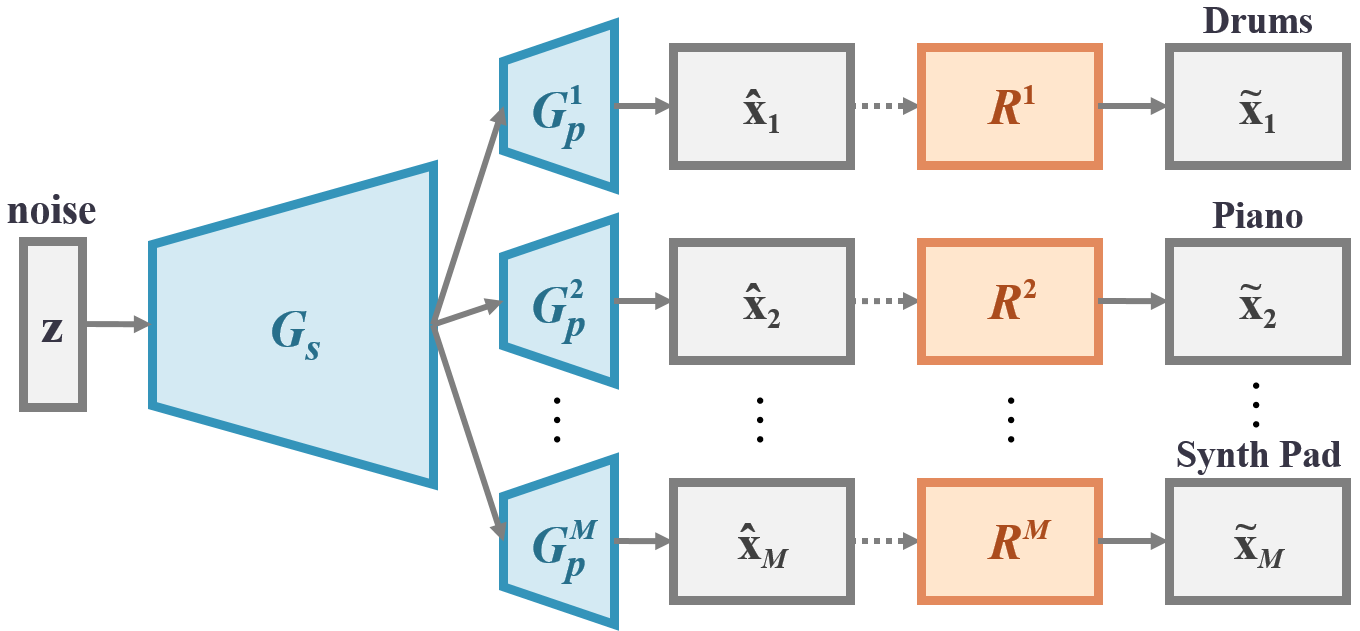}
\caption{The generator and the refiner. The generator ($G_s$ and several $G_p^i$ collectively) produces real-valued predictions. The refiner network (several $R^i$) refines the outputs of the generator into binary ones.}
\label{fig:generator}
\end{figure}

\begin{figure}[t]
\centering
\includegraphics[width=\linewidth]{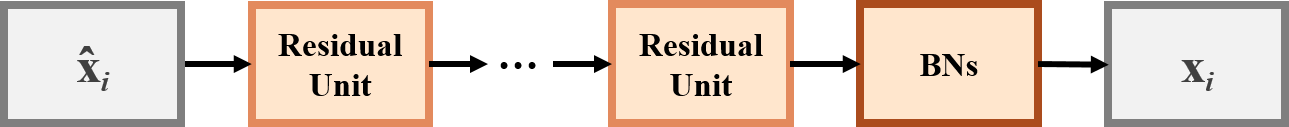}
\caption{The refiner network. The tensor size remains the same throughout the network.}
\label{fig:refiner}
\end{figure}

\subsection{Straight-through Estimator}

Computing the exact gradients for either DBNs or SBNs, however, is intractable. For SBNs, it requires the computation of the average loss over all possible binary samplings of all the SBNs, which is exponential in the total number of SBNs. For DBNs, the threshold function in Eq.~(\ref{eq:dbn}) is non-differentiable. Therefore, the flow of backpropagation used to train parameters of the network would be blocked. 

A few solutions have been proposed to address this issue~\cite{r2rt,bengio}. One strategy is to replace the non-differentiable functions, which are used in the forward pass, by differentiable functions (usually called the \emph{estimators}) in the backward pass. An example is the \emph{straight-through} (ST) estimator proposed by Hinton~\cite{hinton}. In the backward pass, ST simply treats BNs as identify functions and ignores their gradients. A variant of the ST estimator is the \emph{sigmoid-adjusted ST estimator}~\cite{hmrnn}, which multiplies the gradients in the backward pass by the derivative of the sigmoid function. Such estimators were originally proposed as regularizers~\cite{hinton} and later found promising for conditional computation~\cite{bengio}. We use the sigmoid-adjusted ST estimator in training neural networks with BNs and found it empirically works well for our generation task as well.

\section{Proposed Model}
\label{sec:model}

\subsection{Data Representation}
\label{sec:representation}

Following~\cite{musegan}, we use the \textit{multi-track piano-roll} representation. A multi-track piano-roll is defined as a set of piano-rolls for different tracks (or instruments). Each piano-roll is a binary-valued score-like matrix, where its vertical and horizontal axes represent note pitch and time, respectively. The values indicate the presence of notes over different time steps. For the temporal axis, we discard the tempo information and therefore every beat has the same length regardless of tempo.

\begin{figure}[t]
\centering
\includegraphics[width=\linewidth]{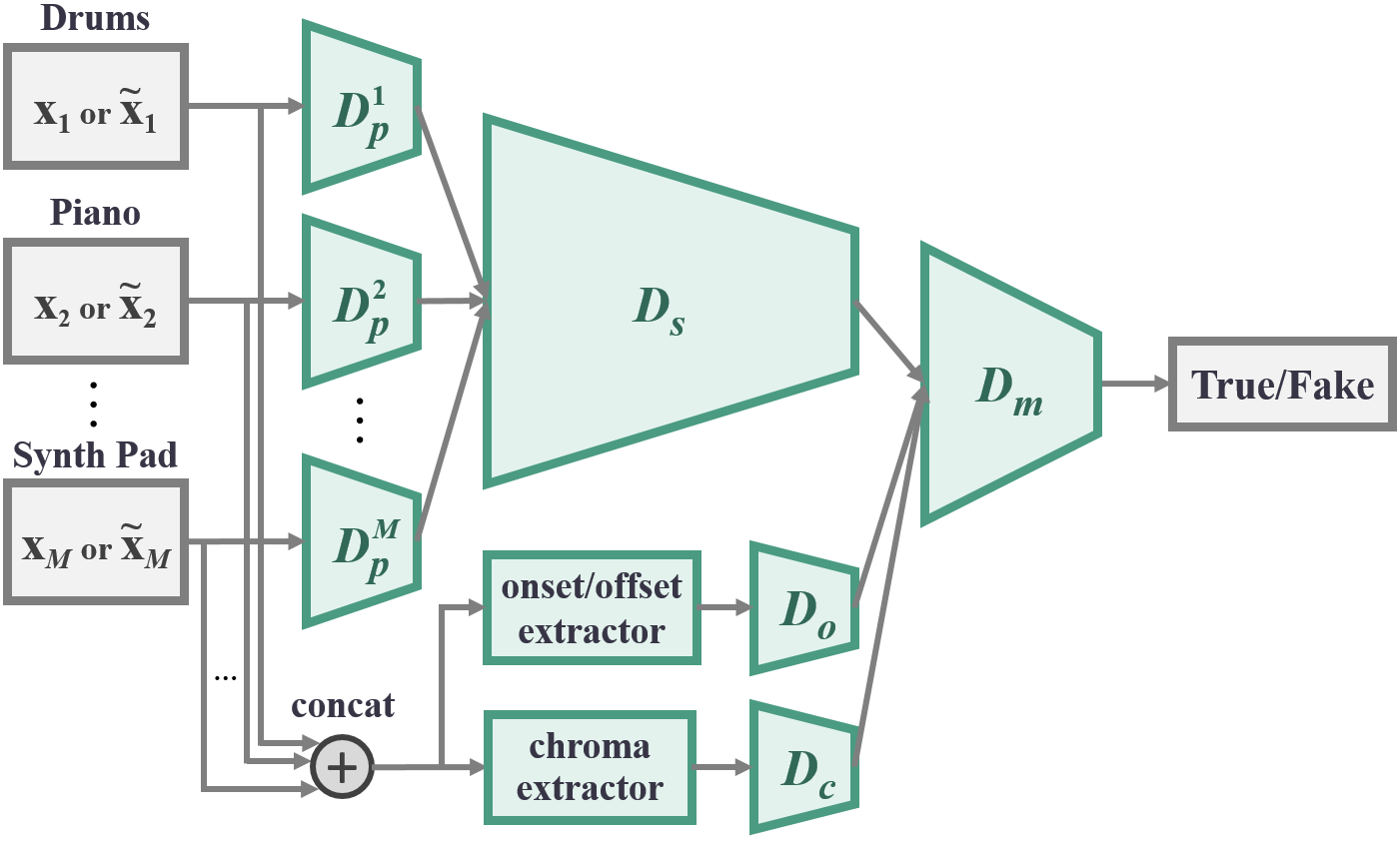}
\caption{The discriminator. It consists of three streams: the main stream ($D_m$, $D_s$ and several $D_p^i$; the upper half), the onset/offset stream ($D_o$) and the chroma stream ($D_c$).}
\label{fig:discriminator}
\end{figure}

\subsection{Generator}
\label{subsec:generator}

As \figref{fig:generator} shows, the generator $G$ consists of a ``s''hared network $G_s$ followed by $M$ ``p''rivate network $G_p^i$, $i=1, \dots, M$, one for each track. The shared generator $G_s$ first produces a high-level representation of the output musical segments that is shared by all the tracks. Each private generator $G_p^i$ then turns such abstraction into the final piano-roll output for the corresponding track. The intuition is that different tracks have their own musical properties (e.g., textures, common-used patterns), while jointly they follow a common, high-level musical idea. The design is different from~\cite{musegan} in that the latter does not include a shared $G_s$ in early layers.

\begin{figure}[t]
\centering
\includegraphics[width=0.75\linewidth]{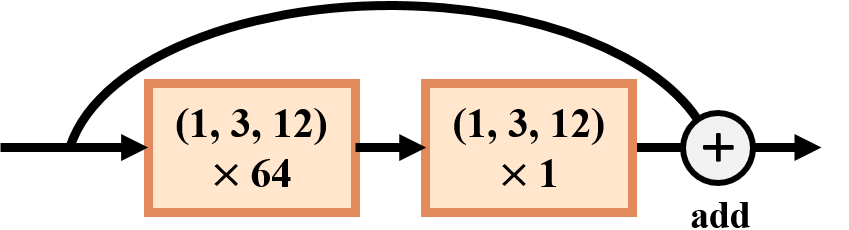}
\caption{Residual unit used in the refiner network. The values denote the kernel size and the number of the output channels of the two convolutional layers.}
\label{fig:resblock}
\end{figure}

\begin{figure*}[t]
\centering
\begin{tabular}{@{}c@{\hspace{1ex}}c@{\hspace{1.5pt}}c@{\hspace{1.5pt}}c@{\hspace{1.5pt}}c@{\hspace{1.5pt}}c@{}}
\shortstack{Dr.\\[10pt]Pi.\\[10pt]Gu.\\[10pt]Ba.\\[10pt]En.\\[4pt]}
&\includegraphics[width=0.19\linewidth]{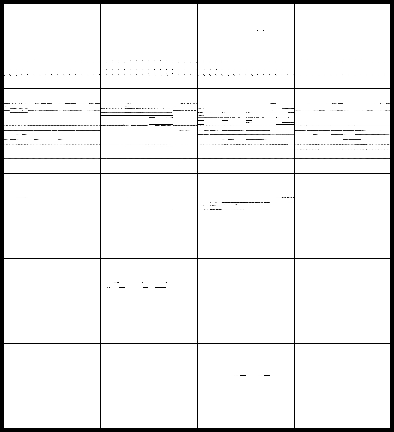}
&\includegraphics[width=0.19\linewidth]{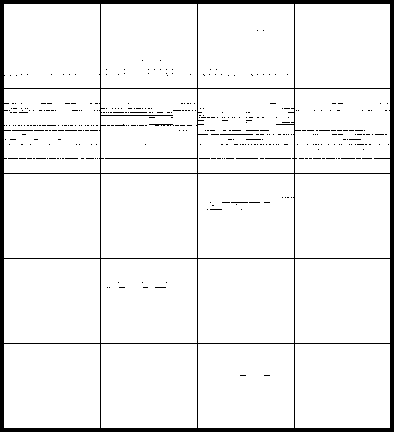}
&\includegraphics[width=0.19\linewidth]{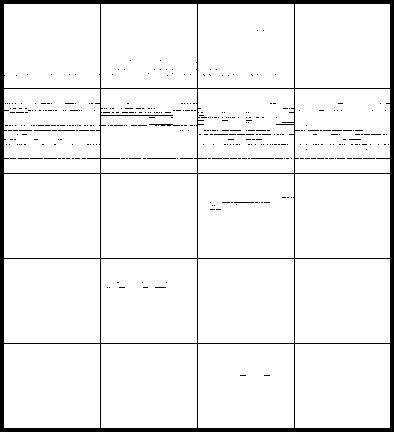}
&\includegraphics[width=0.19\linewidth]{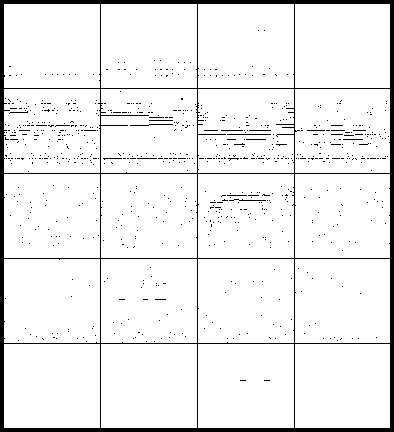}
&\includegraphics[width=0.19\linewidth]{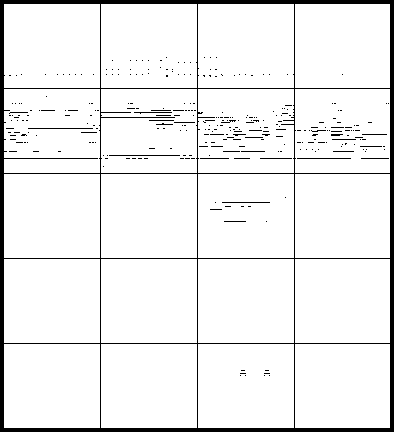}\\
&(a) raw predictions &(b) pretrained (+BS) &(c) pretrained (+HT) &(d) proposed (+SBNs) &(d) proposed (+DBNs)
\end{tabular}
\caption[Caption for LOF]{Comparison of binarization strategies. (a): the probabilistic, real-valued (raw) predictions of the pretrained $G$. (b), (c): the results of applying post-processing algorithms directly to the raw predictions in (a). (d), (e): the results of the proposed models, using an additional refiner $R$ to binarize the real-valued predictions of $G$. Empty tracks are not shown. (We note that in (d), few noises (33 pixels) occur in the \textit{Reed} and \textit{Synth Lead} tracks.)}
\label{fig:binarization_strategies}
\end{figure*}

\subsection{Refiner}
\label{subsec:refiner}

The refiner $R$ is composed of $M$ private networks $R^i$, $i=1, \dots, M$, again one for each track. The refiner aims to refine the real-valued outputs of the generator, $\xh=G(\z)$, into binary ones, $\mathbf{\tilde{x}}$, rather than learning a new mapping from $G(\z)$ to the data space. Hence, we draw inspiration from \emph{residual learning} and propose to construct the refiner with a number of \emph{residual units}~\cite{resunit}, as shown in \figref{fig:refiner}. The output layer (i.e. the final layer) of the refiner is made up of either DBNs or SBNs.

\subsection{Discriminator}
\label{subsec:discriminator}

Similar to the generator, the discriminator $D$ consists of $M$ private network $D_p^i$, $i=1, \dots, M$, one for each track, followed by a shared network $D_s$, as shown in \figref{fig:discriminator}. Each private network $D_p^i$ first extracts low-level features from the corresponding track of the input piano-roll. Their outputs are concatenated and sent to the shared network $D_s$ to extract higher-level abstraction shared by all the tracks. The design differs from~\cite{musegan} in that only one (shared) discriminator was used in~\cite{musegan} to evaluate all the tracks collectively. We intend to evaluate such a new shared/private design in \secref{sec:discriminator_design_exp}.

As a minor contribution, to help the discriminator extract musically-relevant features, we propose to add to the discriminator two more streams, shown in the lower half of \figref{fig:discriminator}. In the first \emph{onset/offset stream}, the differences between adjacent elements in the piano-roll along the time axis are first computed, and then the resulting matrix is summed along the pitch axis, which is finally fed to $D_o$.

In the second \emph{chroma stream}, the piano-roll is viewed as a sequence of one-beat-long frames. A chroma vector is then computed for each frame and jointly form a matrix, which is then be fed to $D_c$. Note that all the operations involved in computing the chroma and onset/offset features are differentiable, and thereby we can still train the whole network by backpropagation.

Finally, the features extracted from the three streams are concatenated and fed to $D_m$ to make the final prediction.

\subsection{Training}

We propose to train the model in a \textbf{two-stage} manner: $G$ and $D$ are pretrained in the first stage; $R$ is then trained along with $D$ (fixing $G$) in the second stage. Other training strategies are discussed and compared in Section \ref{sec:training_strategies}.

\section{Analysis of the Generated Results}
\label{experiment}

\subsection{Training Data \& Implementation Details}

The Lakh Pianoroll Dataset (LPD)~\cite{musegan}\footnote{\url{https://salu133445.github.io/lakh-pianoroll-dataset/}} contains 174,154 multi-track piano-rolls derived from the MIDI files in the Lakh MIDI Dataset (LMD)~\cite{raffel16phd}.\footnote{\url{http://colinraffel.com/projects/lmd/}} In this paper, we use a cleansed subset (\textit{LPD-cleansed}) as the training data, which contains 21,425 multi-track piano-rolls that are in 4/4 time and have been matched to distinct entries in Million Song Dataset (MSD)~\cite{msd}. To make the training data cleaner, we consider only songs with an \textit{alternative} tag. We randomly pick six four-bar phrases from each song, which leads to the final training set of 13,746 phrases from 2,291 songs.

We set the temporal resolution to 24 time steps per beat to cover common temporal patterns such as triplets and 32th notes. An additional one-time-step-long pause is added between two consecutive (i.e. without a pause) notes of the same pitch to distinguish them from one single note. The note pitch has 84 possibilities, from \texttt{C1} to \texttt{B7}.

We categorize all instruments into drums and sixteen instrument families according to the specification of General MIDI Level 1.\footnote{\url{https://www.midi.org/specifications/item/gm-level-1-sound-set}} We discard the less popular instrument families in LPD and use the following eight tracks: \textit{Drums, Piano, Guitar, Bass, Ensemble, Reed, Synth Lead} and \textit{Synth Pad}. Hence, the size of the target output tensor is $4$ (bar) $\times$ $96$ (time step) $\times$ $84$ (pitch) $\times$ $8$ (track).

\begin{table*}[t]
\setlength{\tabcolsep}{6.9pt}
\centering
\begin{tabular}{cccccccccccccc}
\toprule
&\multirow{2}{*}{\raisebox{-3ex}{\shortstack[c]{training\\data}}} &\multicolumn{2}{c}{pretrained} &\multicolumn{2}{c}{proposed} &\multicolumn{2}{c}{joint} &\multicolumn{2}{c}{end-to-end} &\multicolumn{2}{c}{ablated-I} &\multicolumn{2}{c}{ablated-II}\\
\cmidrule(lr){3-4} \cmidrule(lr){5-6} \cmidrule(lr){7-8} \cmidrule(lr){9-10} \cmidrule(lr){11-12} \cmidrule(lr){13-14}
&&BS &HT &SBNs &DBNs &SBNs &DBNs &SBNs &DBNs &BS &HT &BS &HT\\
\midrule
\textbf{QN} &0.88 &\textbf{0.67} &\textbf{0.72} &0.42 &\underline{\textbf{0.78}} &0.18 &0.55 &\textbf{0.67} &0.28 &0.61 &0.64 &0.35 &0.37\\
\textbf{PP} &0.48 &0.20 &0.22 &\textbf{0.26} &\underline{\textbf{0.45}} &0.19 &0.19 &0.16 &\textbf{0.29} &0.19 &0.20 &0.14 &0.14\\
\textbf{TD} &0.96 &\textbf{0.98} &1.00 &\textbf{0.99} &0.87 &\underline{\textbf{0.95}} &1.00 &1.40 &1.10 &1.00 &1.00 &1.30 &1.40\\
\bottomrule
\end{tabular}\\[1pt]
{\footnotesize (Underlined and bold font indicate respectively the top and top-three entries with values closest to those shown in the `training data' column.)}
\caption{Evaluation results for different models. Values closer to those reported in the `training data' column are better.}
\label{tab:score}
\end{table*}

Both $G$ and $D$ are implemented as deep CNNs (see \appref{app:sec:arch} for the detailed network architectures). The length of the input random vector is $128$. $R$ consists of two residual units~\cite{resunit} shown in \figref{fig:resblock}. Following~\cite{wgan-gp}, we use the Adam optimizer~\cite{adam} and only apply batch normalization to $G$ and $R$. We apply the \textit{slope annealing trick}~\cite{hmrnn} to networks with BNs, where the slope of the sigmoid function in the sigmoid-adjusted ST estimator is multiplied by $1.1$ after each epoch. The batch size is $16$ except for the first stage in the two-stage training setting, where the batch size is $32$.

\begin{figure}[t]
\centering
\begin{tabular}{@{}c@{\hspace{1ex}}c@{}}
\raisebox{.6cm}{(a)} &\includegraphics[width=.92\linewidth]{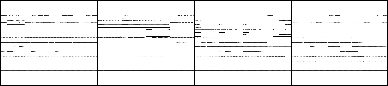}\\[-2pt]
\raisebox{.6cm}{(b)} &\includegraphics[width=.92\linewidth]{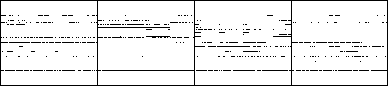}\\[-2pt]
\raisebox{.6cm}{(c)} &\includegraphics[width=.92\linewidth]{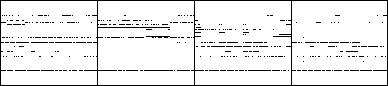}\\[-2pt]
\raisebox{.6cm}{(d)} &\includegraphics[width=.92\linewidth]{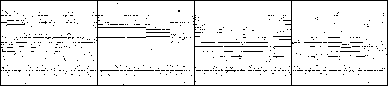}\\[-2pt]
\raisebox{.6cm}{(e)} &\includegraphics[width=.92\linewidth]{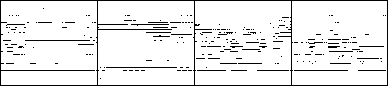}
\end{tabular}
\caption{Closeup of the piano track in \figref{fig:binarization_strategies}.}
\label{fig:closeup}
\end{figure}

\subsection{Objective Evaluation Metrics}

We generate 800 samples for each model (see \appref{app:sec:sample_results} for sample generated results) and use the following metrics proposed in~\cite{musegan} for evaluation. We consider a model better if the average metric values of the generated samples are closer to those computed from the training data.
\begin{itemize}
  \item \textbf{Qualified note rate} (QN) computes the ratio of the number of the qualified notes (notes no shorter than three time steps, i.e., a 32th note) to the total number of notes. Low QN implies overly-fragmented music.
  \item \textbf{Polyphonicity} (PP) is defined as the ratio of the number of time steps where more than two pitches are played to the total number of time steps.
  \item \textbf{Tonal distance} (TD) measures the distance between the chroma features (one for each beat) of a pair of tracks in the tonal space proposed in~\cite{tonaldist}. In what follows, we only report the \textbf{TD} between the piano and the guitar, for they are the two most used tracks.
\end{itemize}

\subsection{Comparison of Binarization Strategies}

We compare the proposed model with two common test-time binarization strategies: \textit{Bernoulli sampling} (BS) and \textit{hard thresholding} (HT). Some qualitative results are provided in Figures \ref{fig:binarization_strategies} and \ref{fig:closeup}. Moreover, we present in \tabref{tab:score} a quantitative comparison among them.

\begin{figure}[t]
\centering
(a) \includegraphics[trim={0 0.27cm 0 0.2cm}, clip, width=0.925\linewidth]{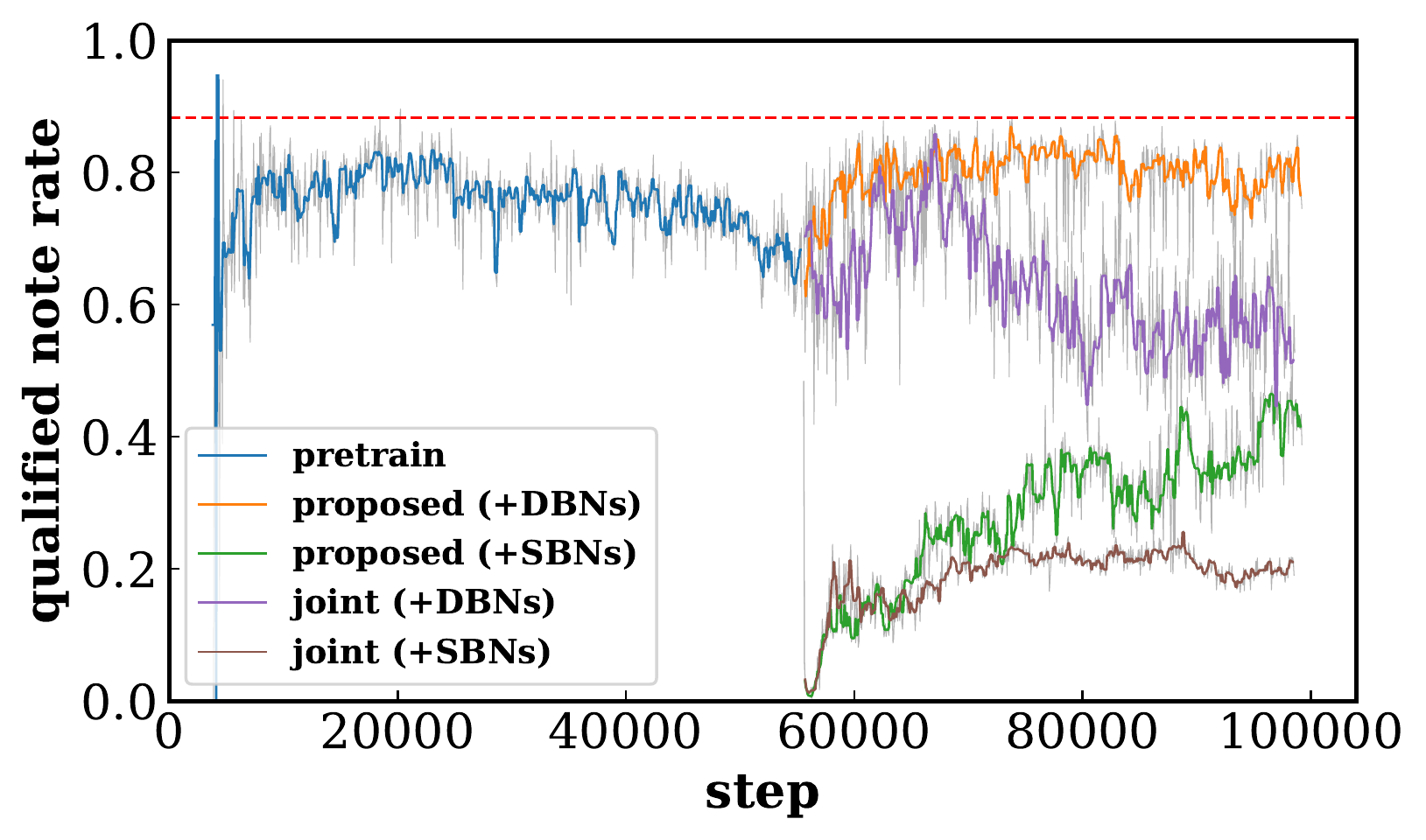}\\[1ex]
(b) \includegraphics[trim={0 0.27cm 0 0.2cm}, clip, width=0.92\linewidth]{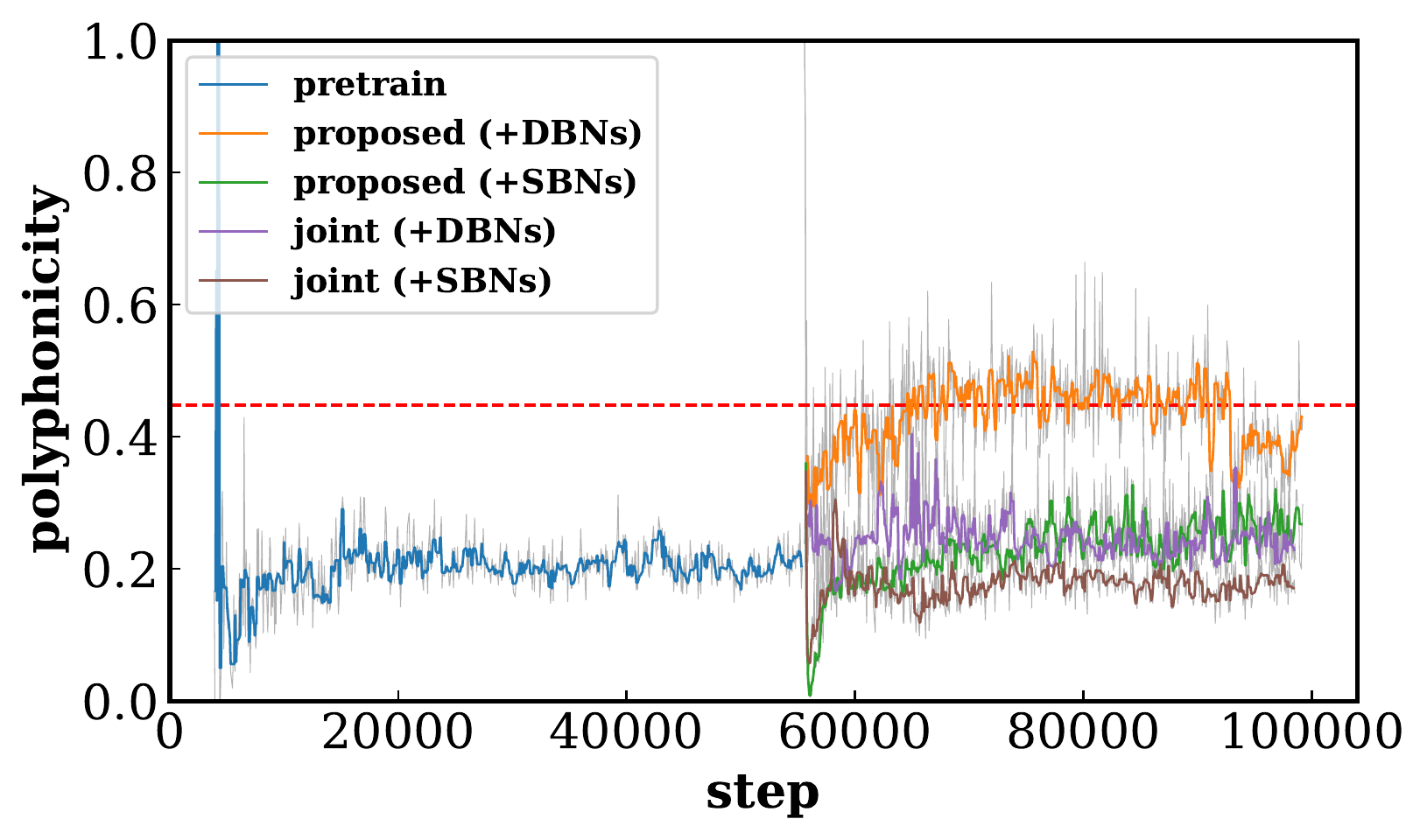}
\caption{(a) Qualified note rate (QN) and (b) polyphonicity (PP) as a function of training steps for different models. The dashed lines indicate the average QN and PP of the training data, respectively. (Best viewed in color.)}
\label{fig:exp1}
\end{figure}

Both qualitative and quantitative results show that the two test-time binarization strategies can lead to overly-fragmented piano-rolls (see the ``pretrained'' ones). The proposed model with DBNs is able to generate piano-rolls with a relatively small number of overly-fragmented notes (a \textbf{QN} of $0.78$; see \tabref{tab:score}) and to better capture the statistical properties of the training data in terms of \textbf{PP}. However, the proposed model with SBNs produces a number of random-noise-like artifacts in the generated piano-rolls, as can be seen in \figref{fig:closeup}(d), leading to a low \textbf{QN} of $0.42$. We attribute to the stochastic nature of SBNs. Moreover, we can also see from \figref{fig:exp1} that only the proposed model with DBNs keeps improving after the second-stage training starts in terms of \textbf{QN} and \textbf{PP}.

\subsection{Comparison of Training Strategies}
\label{sec:training_strategies}

We consider two alternative training strategies:
\begin{itemize}
  \item \textbf{joint}: pretrain $G$ and $D$ in the first stage, and then train $G$ and $R$ (like viewing $R$ as part of $G$) jointly with $D$ in the second stage.
  \item \textbf{end-to-end}: train $G$, $R$ and $D$ jointly in one stage.
\end{itemize}

As shown in \tabref{tab:score}, the models with DBNs trained using the \textit{joint} and \textit{end-to-end} training strategies receive lower scores as compared to the \textit{two-stage} training strategy in terms of \textbf{QN} and \textbf{PP}. We can also see from \figref{fig:exp1}(a) that the model with DBNs trained using the \textit{joint} training strategy starts to degenerate in terms of \textbf{QN} at about 10,000 steps after the second-stage training begins.

\begin{figure}[t]
\centering
\begin{minipage}{0.05\linewidth}
  Dr. \\[3.2pt]
  Pi. \\[3.2pt]
  Gu. \\[3.2pt]
  Ba. \\[3.2pt]
  En. \\[3.2pt]
  Re. \\[3.2pt]
  S.L.\\[3.2pt]
  S.P.\\[5.5pt]
  Dr. \\[3.2pt]
  Pi. \\[3.2pt]
  Gu. \\[3.2pt]
  Ba. \\[3.2pt]
  En.
\end{minipage}
\hspace*{\fill}
\begin{minipage}{0.91\linewidth}
  \includegraphics[width=\linewidth]{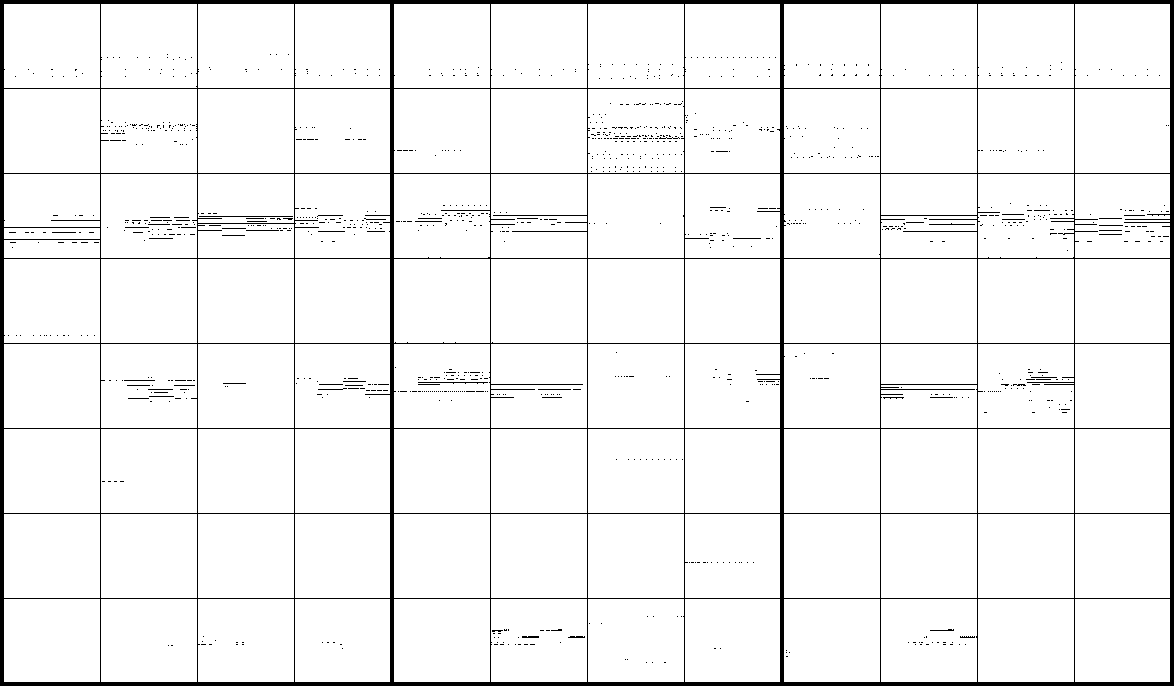}\\[1pt]
  \includegraphics[width=\linewidth]{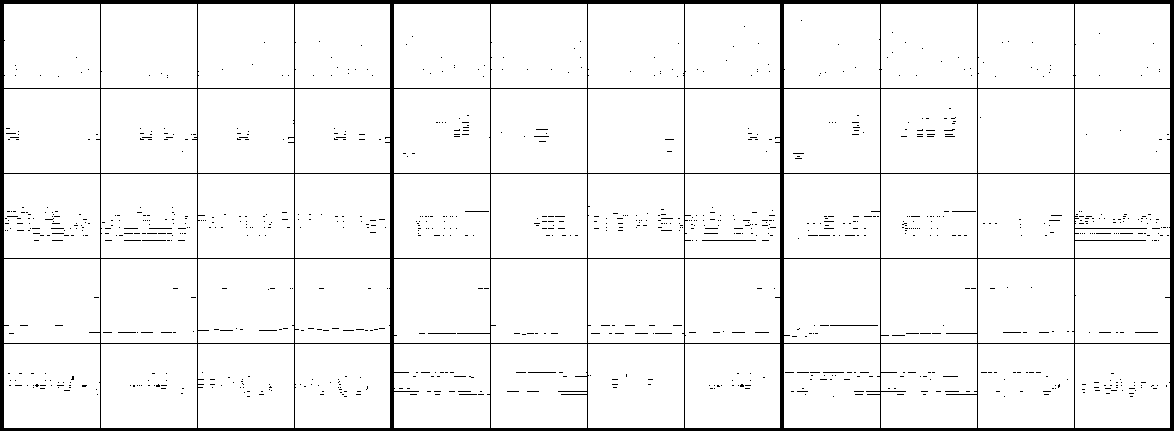}
\end{minipage}
\caption{Example generated piano-rolls of the end-to-end models with (top) DBNs and (bottom) SBNs. Empty tracks are not shown.}
\label{fig:end2end}
\end{figure}

\figref{fig:end2end} shows some qualitative results for the \textit{end-to-end} models. It seems that the models learn the proper pitch ranges for different tracks. We also see some chord-like patterns in the generated piano-rolls. From \tabref{tab:score} and \figref{fig:end2end}, in the end-to-end training setting SBNs are not inferior to DBNs, unlike the case in the two-stage training. Although the generated results appear preliminary, to our best knowledge this represents the first attempt to generate such high dimensional data with BNs from scratch (see remarks in \appref{app:sec:end2end}).

\subsection{Effects of the Shared/private and Multi-stream Design of the Discriminator}
\label{sec:discriminator_design_exp}

We compare the proposed model with two ablated versions: the \textbf{ablated-I} model, which removes the onset/offset and chroma streams, and the \textbf{ablated-II} model, which uses only a shared discriminator without the shared/private and multi-stream design (i.e., the one adopted in~\cite{musegan}).\footnote{The number of parameters for the proposed, ablated-I and ablated-II models is 3.7M, 3.4M and 4.6M, respectively.} Note that the comparison is done by applying either BS or HT (not BNs) to the first-stage pretrained models.

\begin{figure}[t]
\centering
\includegraphics[trim={0 0.28cm 0 0.26cm}, clip, width=.95\linewidth]{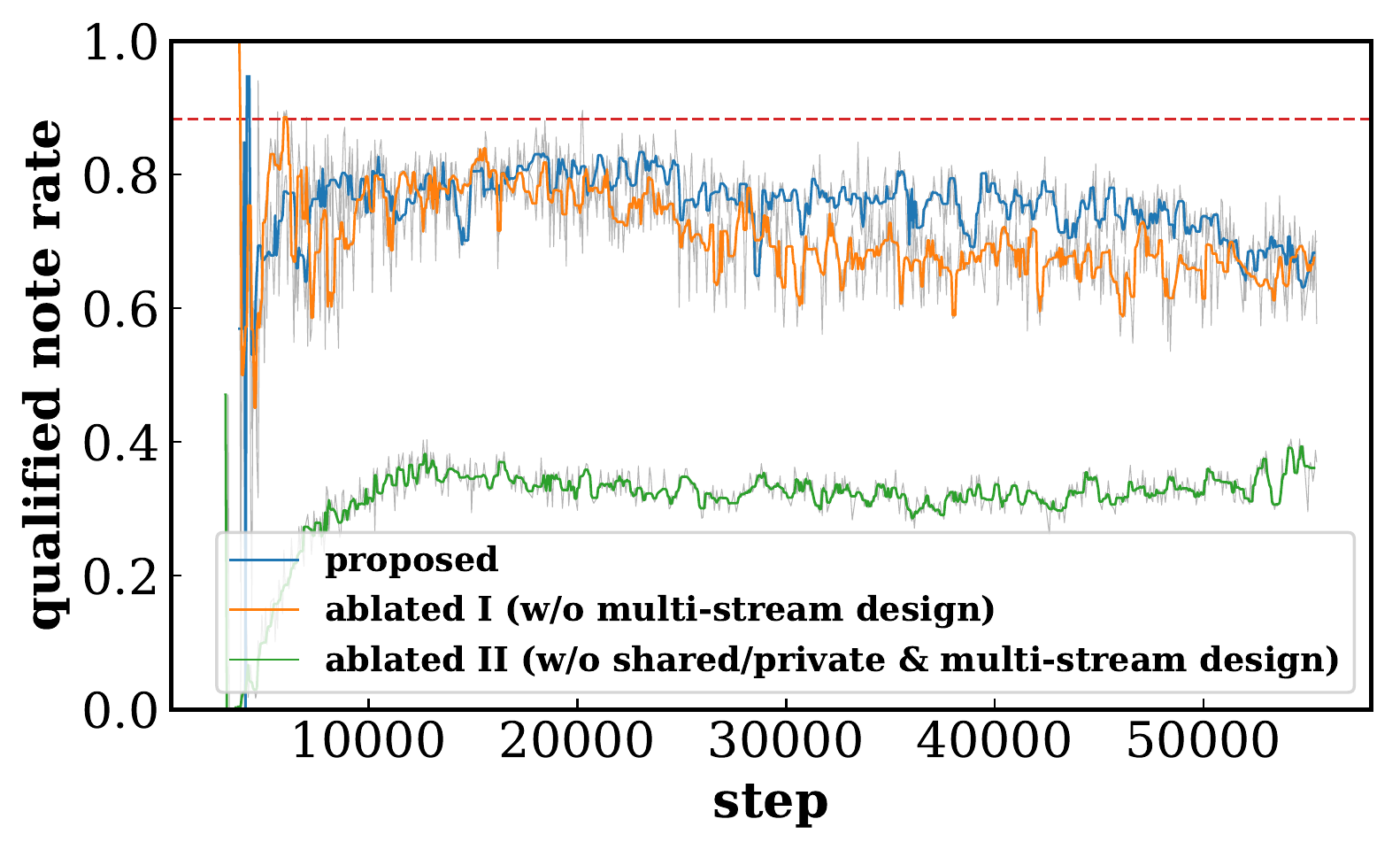}
\caption{Qualified note rate (QN) as a function of training steps for different models. The dashed line indicates the average QN of the training data. (Best viewed in color.)}
\label{fig:exp2}
\end{figure}

As shown in \tabref{tab:score}, the proposed model (see ``pretrained'') outperforms the two ablated versions in all three metrics. A lower \textbf{QN} for the proposed model as compared to the ablated-I model suggests that the onset/offset stream can alleviate the overly-fragmented note problem. Lower \textbf{TD} for the proposed and ablated-I models as compared to the ablated-II model indicates that the shared/private design better capture the intertrack harmonicity. \figref{fig:exp2} also shows that the proposed and ablated-I models learn faster and better than the ablated-II model in terms of \textbf{QN}.

\subsection{User Study}
\label{userstudy}

Finally, we conduct a user study involving 20 participants recruited from the Internet. In each trial, each subject is asked to compare two pieces of four-bar music generated from scratch by the proposed model using SBNs and DBNs, and vote for the better one in four measures. There are five trials in total per subject. We report in \tabref{tab:userstudy} the ratio of votes each model receives. The results show a preference to DBNs for the proposed model.

\begin{table}[t]
\centering
\begin{tabular}{lcc}
\toprule
&with SBNs &with DBNs\\
\midrule
completeness\textsuperscript{*} &0.19 &\textbf{0.81}\\
harmonicity &0.44 &\textbf{0.56}\\
rhythmicity &\textbf{0.56} &0.44\\
overall rating & 0.16 &\textbf{0.84}\\
\bottomrule
\end{tabular}\\
{\footnotesize \textsuperscript{*}We asked, ``Are there many overly-fragmented notes?"}
\caption{Result of a user study, averaged over 20 subjects.}
\label{tab:userstudy}
\end{table}

\section{Discussion and Conclusion}
\label{sec:conclusions}

We have presented a novel convolutional GAN-based model for generating binary-valued piano-rolls by using binary neurons at the output layer of the generator. We trained the model on an eight-track piano-roll dataset. Analysis showed that the generated results of our model with deterministic binary neurons features fewer overly-fragmented notes as compared with existing methods. Though the generated results appear preliminary and lack musicality, we showed the potential of adopting binary neurons in a music generation system. 

In future work, we plan to further explore the end-to-end models and add recurrent layers to the temporal model. It might also be interesting to use BNs for music transcription \cite{Benetos2013}, where the desired outputs are also binary-valued.

\bibliography{ref}

\newpage

\begin{appendices}
\setcounter{secnumdepth}{1}

\begin{center}
\textbf{\Large APPENDIX}
\end{center}

\section{Network Architectures}
\label{app:sec:arch}
We show in \tabref{app:tab:arch} the network architectures for the generator $G$, the discriminator $D$, the onset/offset feature extractor $D_o$, the chroma feature extractor $D_c$ and the discriminator for the ablated-II model.

\section{Samples of the Training Data}
\label{app:sec:sample_train}

\begin{figure*}[t]
\centering
\includegraphics[scale=0.24]{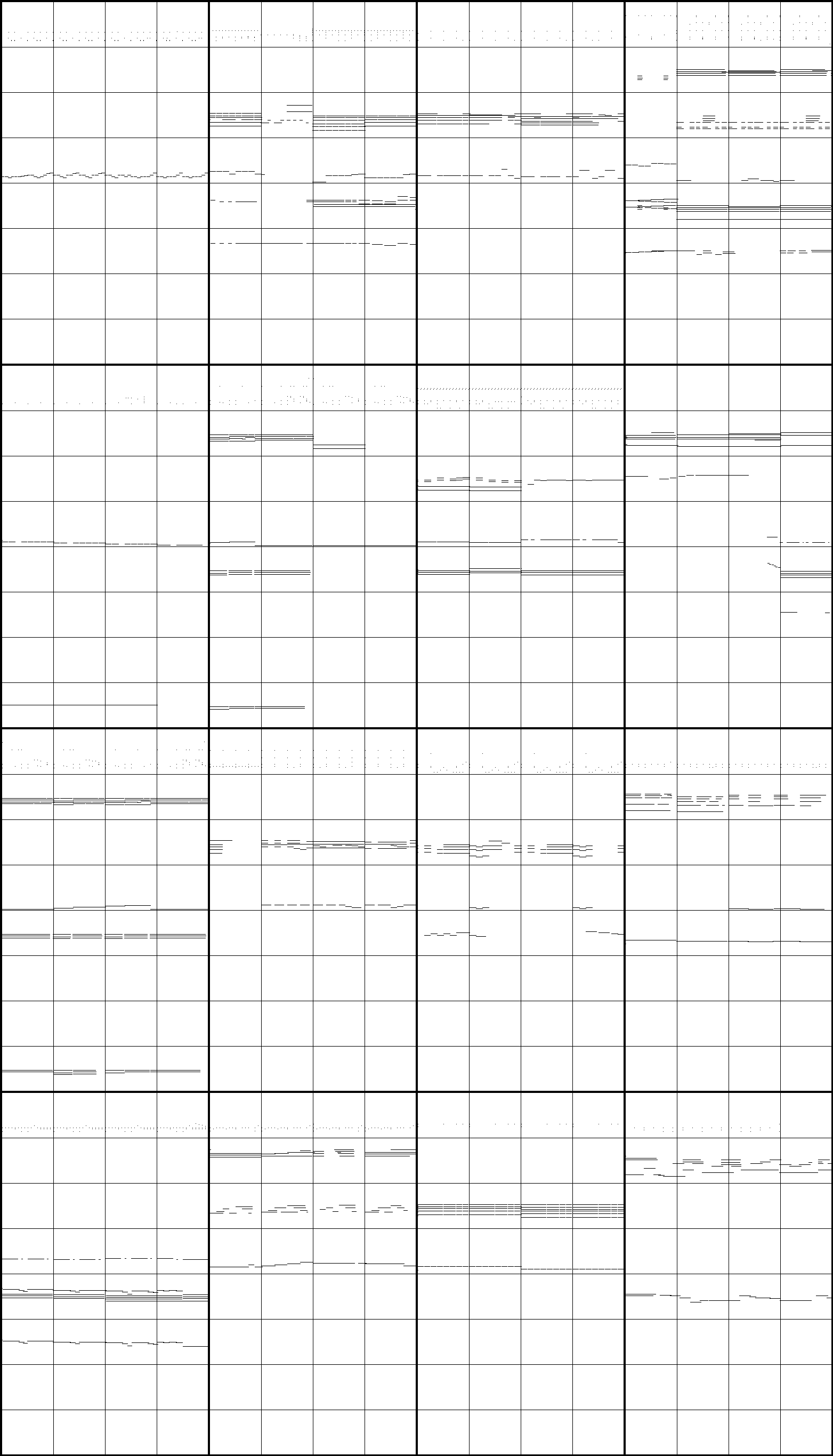}
\caption{Sample eight-track piano-rolls seen in the training data. Each block represents a bar for a certain track. The eight tracks are (from top to bottom) \textit{Drums, Piano, Guitar, Bass, Ensemble, Reed, Synth Lead} and \textit{Synth Pad}.}
\label{app:fig:sample_train}
\end{figure*}

\figref{app:fig:sample_train} shows some sample eight-track piano-rolls seen in the training data.

\section{Sample Generated Piano-rolls}
\label{app:sec:sample_results}

\begin{figure*}[t]
\centering
\includegraphics[scale=0.24]{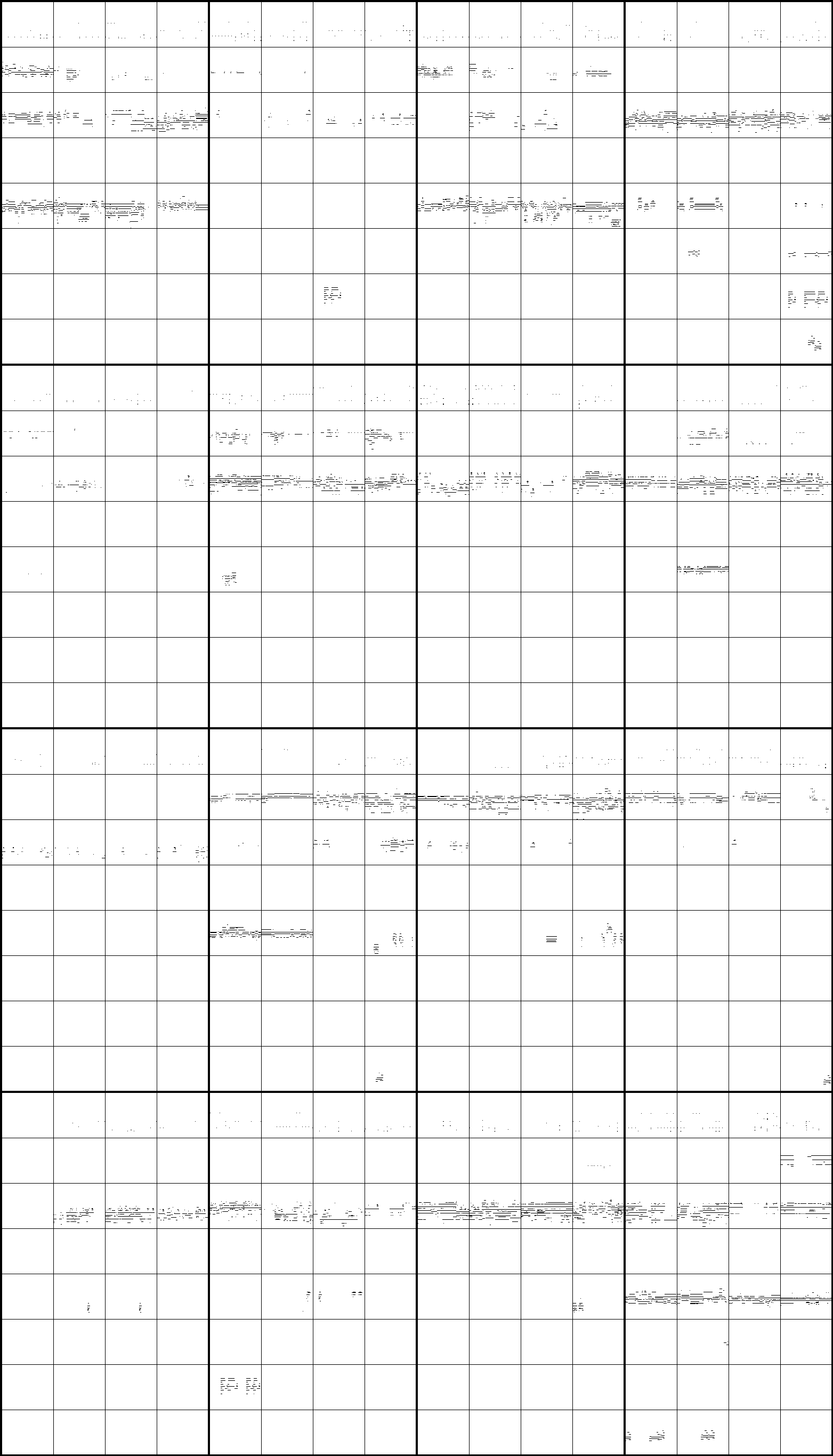}
\caption{Randomly-chosen eight-track piano-rolls generated by the proposed model with DBNs. Each block represents a bar for a certain track. The eight tracks are (from top to bottom) \textit{Drums, Piano, Guitar, Bass, Ensemble, Reed, Synth Lead} and \textit{Synth Pad}.}
\label{app:fig:sample_dbn}
\end{figure*}

\begin{figure*}[t]
\centering
\includegraphics[scale=0.24]{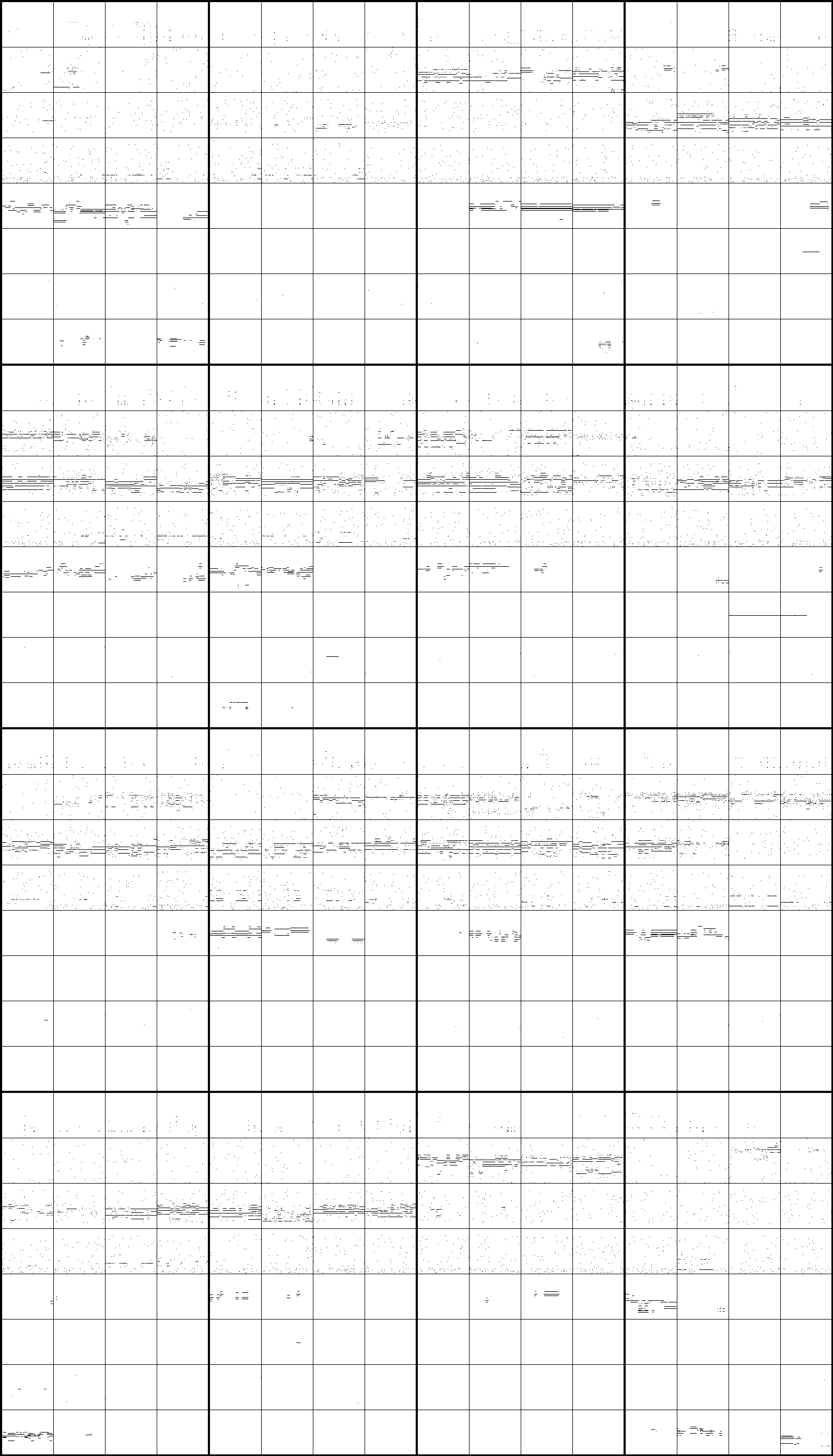}
\caption{Randomly-chosen eight-track piano-rolls generated by the proposed model with SBNs. Each block represents a bar for a certain track. The eight tracks are (from top to bottom) \textit{Drums, Piano, Guitar, Bass, Ensemble, Reed, Synth Lead} and \textit{Synth Pad}.}
\label{app:fig:sample_sbn}
\end{figure*}

We show in Figures~\ref{app:fig:sample_dbn} and \ref{app:fig:sample_sbn} some sample eight-track piano-rolls generated by the proposed model with DBNs and SBNs, respectively.

\section{Remarks on the End-to-end Models}
\label{app:sec:end2end}

\begin{figure*}[t]
\centering
\includegraphics[width=\linewidth]{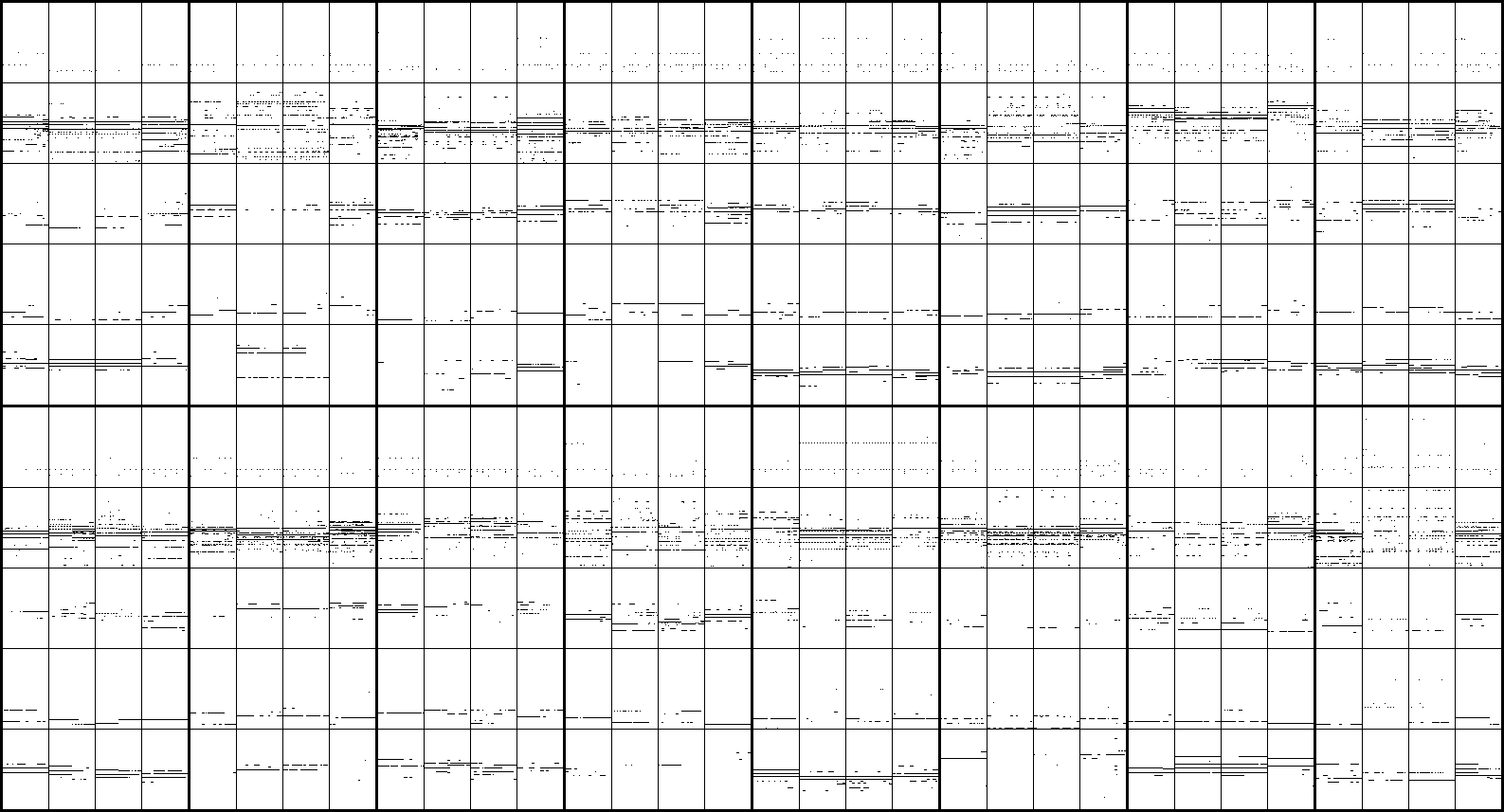}\\
(a) modified end-to-end model (+DBNs)\\[1em]
\includegraphics[width=\linewidth]{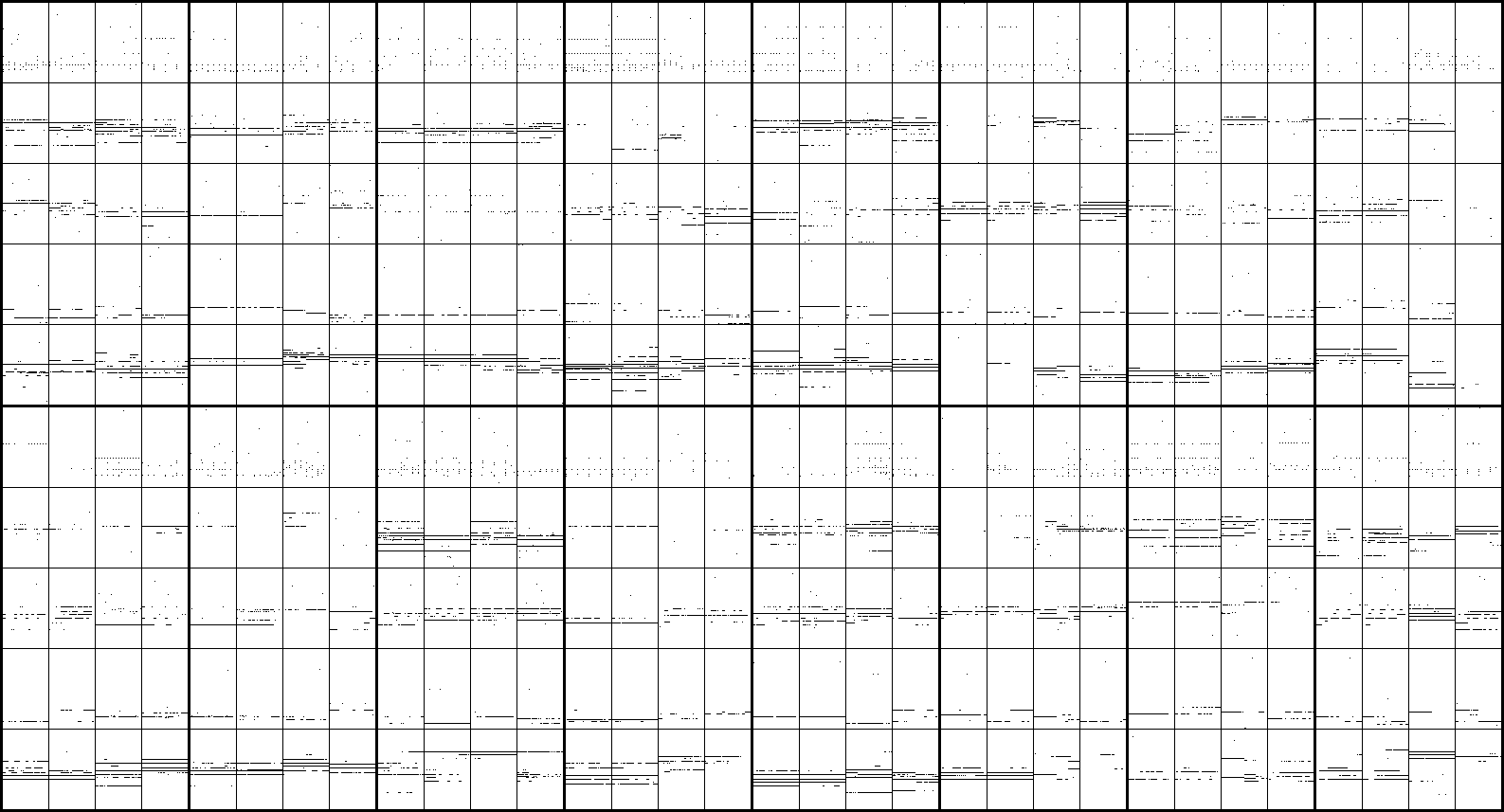}\\
(b) modified end-to-end model (+SBNs)
\caption{Randomly-chosen five-track piano-rolls generated by the modified end-to-end models (see \appref{app:sec:end2end}) with (a) DBNs and (b) SBNs. Each block represents a bar for a certain track. The five tracks are (from top to bottom) \textit{Drums, Piano, Guitar, Bass} and \textit{Ensemble}.}
\label{app:fig:sample_end2end}
\end{figure*}

After several trials, we found that the claim in the main text that an end-to-end training strategy cannot work well is not true with the following modifications to the network. However, a thorough analysis of the end-to-end models are beyond the scope of this paper.
\begin{itemize}
  \item remove the refiner network ($R$)
  \item use binary neurons (either DBNs or SBNs) in the last layer of the generator ($G$) 
  \item reduce the temporal resolution by half to $12$ time steps per beat
  \item use five-track (\textit{Drums, Piano, Guitar, Bass} and \textit{Ensemble}) piano-rolls as the training data 
\end{itemize}
We show in \figref{app:fig:sample_end2end} some sample five-track piano-rolls generated by the modified end-to-end models with DBNs and SBNs.

\begin{table*}[t]
\centering
\begin{tabular}{|l|ccccccc|}
  \hline
  \multicolumn{8}{|l|}{\textbf{Input}: $\Re^{128}$}\\
  \hline\hline
  \textit{dense} &\multicolumn{7}{l|}{$1536$}\\
  \hline
  \multicolumn{8}{|c|}{\textit{reshape to $(3, 1, 1)\times512$ channels}}\\
  \hline
  \textit{transconv} &$256$ &$2\times1\times1$ &\multicolumn{1}{c}{$(1, 1, 1)$} &&&&\\
  \textit{transconv} &$128$ &$1\times4\times1$ &\multicolumn{1}{c}{$(1, 4, 1)$} &&&&\\
  \textit{transconv} &$128$ &$1\times1\times3$ &\multicolumn{1}{c}{$(1, 1, 3)$} &&&&\\
  \textit{transconv} &$64$  &$1\times4\times1$ &\multicolumn{1}{c}{$(1, 4, 1)$} &&&&\\
  \textit{transconv} &$64$  &$1\times1\times3$ &\multicolumn{1}{c}{$(1, 1, 2)$} &&&&\\
  \hline
  &\multicolumn{3}{c}{substream I} &\multicolumn{3}{|c|}{substream II} &\multicolumn{1}{c|}{\multirow{5}{*}{$\cdots\times8$}}\\
  \cdashline{2-7}[1pt/1pt]
  \textit{transconv} &$64$ &$1\times1\times12$ &\multicolumn{1}{c|}{$(1, 1, 12)$} &$64$ &$1\times6\times1$  &\multicolumn{1}{c|}{$(1, 6, 1)$}  &\\
  \textit{transconv} &$32$ &$1\times6\times1$  &\multicolumn{1}{c|}{$(1, 6, 1)$}  &$32$ &$1\times1\times12$ &\multicolumn{1}{c|}{$(1, 1, 12)$} &\\
  \cline{1-7}
  \multicolumn{7}{|c|}{\textit{concatenate along the channel axis}} &\\
  \cline{1-7}
  \textit{transconv} &$1$ &$1\times1\times1$  &\multicolumn{1}{c}{$(1, 1, 1)$} &\multicolumn{3}{c|}{} &\\
  \hline
  \multicolumn{8}{|c|}{\textit{stack along the track axis}}\\
  \hline\hline
  \multicolumn{8}{|l|}{\textbf{Output}: $\Re^{4\times96\times84\times8}$}\\
  \hline
\end{tabular}\\[1ex]
(a) generator $G$\\[2em]

\begin{tabular}{|l|ccccccc|c|c|}
  \hline
  \multicolumn{10}{|l|}{\textbf{Input}: $\Re^{4\times96\times84\times8}$}\\
  \hline\hline
  \multicolumn{8}{|c|}{\textit{split along the track axis}} &\multicolumn{1}{c|}{\multirow{9}{*}{\rotatebox[origin=c]{-90}{chroma stream}}} &\multicolumn{1}{c|}{\multirow{9}{*}{\rotatebox[origin=c]{-90}{onset stream}}}\\
  \cline{1-8}
  &\multicolumn{3}{c}{substream I} &\multicolumn{3}{|c|}{substream II} &\multicolumn{1}{c|}{\multirow{5}{*}{$\cdots\times8$}} &&\\
  \cdashline{2-7}[1pt/1pt]
  \textit{conv} &$32$ &$1\times1\times12$ &\multicolumn{1}{c|}{$(1, 1, 12)$} &$32$ &$1\times6\times1$  &\multicolumn{1}{c|}{$(1, 6, 1)$} &&&\\
  \textit{conv} &$64$ &$1\times6\times1$  &\multicolumn{1}{c|}{$(1, 6, 1)$}  &$64$ &$1\times1\times12$ &\multicolumn{1}{c|}{$(1, 1, 12)$} &&&\\
  \cline{1-7}
  \multicolumn{7}{|c|}{\textit{concatenate along the channel axis}} &&&\\
  \cline{1-7}
  \textit{conv} &$64$ &$1\times1\times1$ &$(1, 1, 1)$ &\multicolumn{3}{c|}{} &&&\\
  \cline{1-8}
  \multicolumn{8}{|c|}{\textit{concatenate along the channel axis}} &&\\
  \cline{1-8}
  \textit{conv} &$128$ &$1\times4\times3$  &\multicolumn{1}{c}{$(1, 4, 2)$} &&&&&& \\
  \textit{conv} &$256$ &$1\times4\times3$  &\multicolumn{1}{c}{$(1, 4, 3)$} &&&&&& \\
  \hline
  \multicolumn{10}{|c|}{\textit{concatenate along the channel axis}}\\
  \hline
  \textit{conv} &$512$ &$2\times1\times1$  &\multicolumn{1}{c}{$(1, 1, 1)$} &\multicolumn{6}{c|}{}\\
  \textit{dense} &$1536$ &\multicolumn{8}{c|}{}\\
  \textit{dense} &$1$ &\multicolumn{8}{c|}{}\\
  \hline\hline
  \multicolumn{10}{|l|}{\textbf{Output}: $\Re$}\\
  \hline
\end{tabular}\\[1ex]
(b) discriminator $D$\\[2em]

\begin{minipage}[c]{.45\linewidth}
\centering
\begin{tabular}{|lccc|}
  \hline
  \multicolumn{4}{|l|}{\textbf{Input}: $\Re^{4\times96\times1\times8}$}\\
  \hline\hline
  \textit{conv} &$32$  &$1\times6\times1$  &$(1, 6, 1)$\\
  \textit{conv} &$64$  &$1\times4\times1$  &$(1, 4, 1)$\\
  \textit{conv} &$128$ &$1\times4\times1$  &$(1, 4, 1)$\\
  \hline\hline
  \multicolumn{4}{|l|}{\textbf{Output}: $\Re^{4\times1\times1\times128}$}\\
  \hline
\end{tabular}\\[1ex]
(c) onset/offset feature extractor $D_o$\\[2em]

\begin{tabular}{|lccc|}
  \hline
  \multicolumn{4}{|l|}{\textbf{Input}: $\Re^{4\times4\times12\times8}$}\\
  \hline\hline
  \textit{conv} &$64$  &$1\times1\times12$ &$(1, 1, 12)$\\
  \textit{conv} &$128$ &$1\times4\times1$  &$(1, 4, 1)$\\
  \hline\hline
  \multicolumn{4}{|l|}{\textbf{Output}: $\Re^{4\times1\times1\times128}$}\\
  \hline
\end{tabular}\\[1ex]
(d) chroma feature extractor $D_c$

\end{minipage}
\begin{minipage}{.45\linewidth}
\centering
\begin{tabular}{|lccc|}
  \hline
  \multicolumn{4}{|l|}{\textbf{Input}: $\Re^{4\times96\times84\times8}$}\\
  \hline\hline
  \textit{conv} &$128$  &$1\times1\times12$ &$(1, 1, 12)$\\
  \textit{conv} &$128$  &$1\times1\times3$  &$(1, 1, 2)$\\
  \textit{conv} &$256$  &$1\times6\times1$  &$(1, 6, 1)$\\
  \textit{conv} &$256$  &$1\times4\times1$  &$(1, 4, 1)$\\
  \textit{conv} &$512$  &$1\times1\times3$  &$(1, 1, 3)$\\
  \textit{conv} &$512$  &$1\times4\times1$  &$(1, 4, 1)$\\
  \textit{conv} &$1024$ &$2\times1\times1$  &$(1, 1, 1)$\\
  \hline
  \multicolumn{4}{|c|}{\textit{flatten to a vector}}\\
  \hline
  \textit{dense} &$1$ &&\\
  \hline\hline
  \multicolumn{4}{|l|}{\textbf{Output}: $\Re$}\\
  \hline
\end{tabular}\\[1ex]
(e) discriminator for the ablated-II model
\end{minipage}
\caption{Network architectures for (a) the generator $G$, (b) the discriminator $D$, (c) the onset/offset feature extractor $D_o$, (d) the chroma feature extractor $D_c$ and (e) the discriminator for the ablated-II model. For the convolutional layers (\textit{conv}) and the transposed convolutional layers (\textit{transconv}), the values represent (from left to right): the number of filters, the kernel size and the strides. For the dense layers (\textit{dense}), the value represents the number of nodes. Each transposed convolutional layer in $G$ is followed by a batch normalization layer and then activated by ReLUs except for the last layer, which is activated by sigmoid functions. The convolutional layers in $D$ are activated by LeakyReLUs except for the last layer, which has no activation function.}
\label{app:tab:arch}
\end{table*}

\end{appendices}

\end{document}